\documentclass[11pt,a4paper]{article}
\usepackage[hyperref]{emnlp2020}
\usepackage{times}
\usepackage{latexsym}

\usepackage{microtype}

\usepackage{amsmath,amsfonts,amssymb}
\usepackage[space]{grffile}
\usepackage{subcaption}
\usepackage{courier}
\usepackage{amsthm}
\usepackage{xcolor}
\usepackage{listings}
\usepackage{epstopdf}
\usepackage{epsfig}
\usepackage{amsmath}
\usepackage{multirow}
\usepackage{bm}
\usepackage{adjustbox}
\usepackage{enumitem}
\usepackage{amssymb}
\usepackage{caption}
\usepackage{framed}
\usepackage{tabularx,ragged2e}
\newcolumntype{C}{>{\Centering\arraybackslash}X} 

\AtBeginDocument{%
  \providecommand\BibTeX{{%
    \normalfont B\kern-0.5em{\scshape i\kern-0.25em b}\kern-0.8em\TeX}}}

\DeclareMathOperator*{\argmax}{\arg\!\max}

\newcommand{\nop}[1]{}

\newcommand{\loss}[0]{{\mathcal{L}}}

\aclfinalcopy

\begin{document}

\title{Unsupervised Natural Language Inference via Decoupled Multimodal Contrastive Learning}

\author{
\centerline{Wanyun Cui$^{\S}$ \; Guangyu Zheng$^{\ddag}$ \; Wei Wang$^{\ddag}$ \;}\\
\centerline{cui.wanyun@sufe.edu.cn,
simonzheng96@gmail.com,weiwang1@fudan.edu.cn}\\
{$^{\S}$Shanghai University of Finance and Economics} \\
{$^{\ddag}$Shanghai Key Laboratory of Data Science, Fudan University} 
}



\maketitle
\begin{abstract}
  We propose to solve the natural language inference problem without any supervision from the inference labels via task-agnostic multimodal pretraining. Although recent studies of multimodal self-supervised learning also represent the linguistic and visual context, their encoders for different modalities are coupled. Thus they cannot incorporate visual information when encoding plain text alone. In this paper, we propose \underline{M}ultimodal \underline{A}ligned \underline{C}ontrastive \underline{D}ecoupled learning (MACD) network. MACD forces the decoupled text encoder to represent the visual information via contrastive learning. Therefore, it embeds visual knowledge even for plain text inference. We conducted comprehensive experiments over plain text inference datasets (i.e. SNLI and STS-B). The unsupervised MACD even outperforms the fully-supervised BiLSTM and BiLSTM+ELMO on STS-B.
\end{abstract}




\section{Introduction}
\label{sec:intro}
Humans are not supervised by the natural language inference (NLI). Supervision is necessary for applications in human-defined domains. For example, humans need the supervision of what is a noun before they do POS tagging, or what is a tiger in Wordnet before they classify an image of tiger in ImageNet. However, for NLI, people are able to entail that \textcircled{a} {\tt A man plays a piano} contradicts \textcircled{b} {\tt A man plays the clarinet for his family} without any supervision from the NLI labels. In this paper, we define such inference as a more general process of establishing associations and inferences between texts, rather than strictly classifying whether two sentences entail or contradict each other.
Inspired by this, we raise the core problem in this paper: {\it Given a pair of natural language sentences, can machines entail their relationship without any supervision from inference labels?}

In his highly acclaimed paper, neuroscientist Moshe Bar claims that {\it ``predictions rely on the existing scripts in memory, which are the result of real as well as of previously imagined experiences''}~\cite{bar2009proactive}. The exemplar theory argues that humans use {\bf similarity} to recognize different objects and make decisions~\cite{tversky1973availability,homa1981limitations}.

Analogy helps humans understand a novel object by linking it to a similar representation existing in memory~\cite{bar2007proactive}. Such linking is facilitated by the object itself and its {\it context}~\cite{bar2004visual}. Context information has been widely applied in self-supervision learning (SSL)~\cite{devlin2018bert,de1994learning,he2019momentum}. Adapting context to NLI is even more straightforward. A simple idea of {\bf constant conjunction} is
that {\it A} causes {\it B} if they are constantly conjoined. Although constant conjunction contradicts ``correlation is not causation'',
modern neuroscience has confirmed that humans use it for reasoning in their mental world~\cite{levy1983temporal}. For example, they found an increase in synaptic efficacy arises from a presynaptic cell's repeated and persistent stimulation of a postsynaptic cell in Hebbian theory~\cite{hebb2005organization}. As to the natural language, the object and its context can be naturally used to determine the inference. For example, \textcircled{a} contradicts \textcircled{b} because they cannot happen simultaneously in the same {\bf context}.

The context representation learned by SSL (e.g. BERT~\cite{devlin2018bert}) has already achieved big success in NLP. From the perspective of context, these models~\cite{devlin2018bert,DBLP:journals/corr/abs-1907-11692} learn the sentence level contextual information (i.e. by next sentence prediction task) and the word level contextual information (i.e. by masked language model task).

Besides linguistic contexts, humans also link other modalities (e.g. visions, voices) to novel inputs~\cite{bar2009proactive}.
Even if the goal is to reason about plain texts, other modalities still help (although they are not provided as inputs)~\cite{kiela2018learning}. For example, if only textual information is used, it is difficult to entail the contradiction between \textcircled{a} and \textcircled{b}. We need the commonsense that a man only has two arms, which cannot play the piano and clarinet simultaneously. This commonsense is hard to obtain from the text. However, if we link the sentences to their {\it visual scenes}, the contradiction is much clearer because the two scenes cannot happen in the same visual context. We think it is necessary to incorporate other modalities for the unsupervised natural language inference.

\begin{figure}[htb]
\vskip 0pt
    \begin{subfigure}[b]{0.23\textwidth}
        \centering
            \includegraphics[scale=0.5]{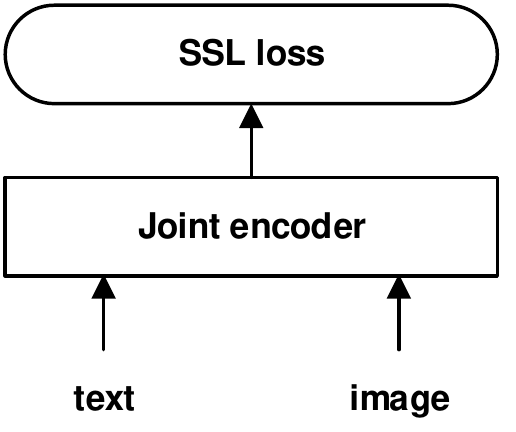}
            \caption{Multimodal SSL with one joint encoder.\newline}
        \label{fig:framework:a}
    \end{subfigure}
    \begin{subfigure}[b]{0.23\textwidth}
        \centering
            \includegraphics[scale=0.5]{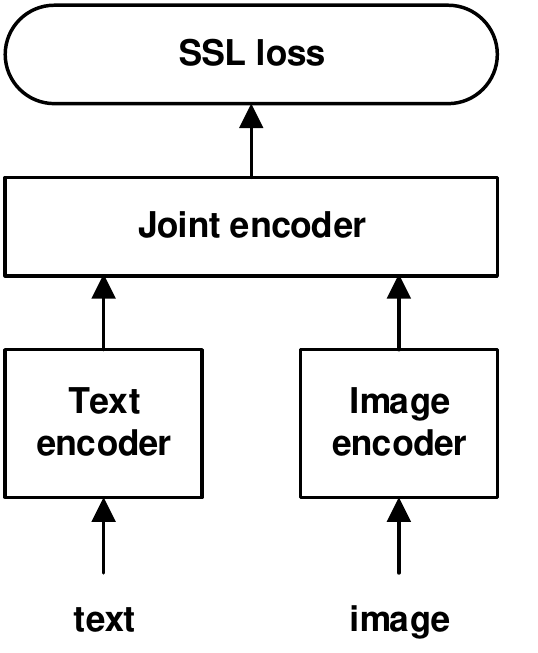}
            \caption{Multimodal SSL with two single-modal encoders and one joint encoder.}
        \label{fig:framework:b}
    \end{subfigure}
    \begin{subfigure}[b]{0.46\textwidth}
        \centering
        \hspace{-1cm}
            \includegraphics[scale=0.5]{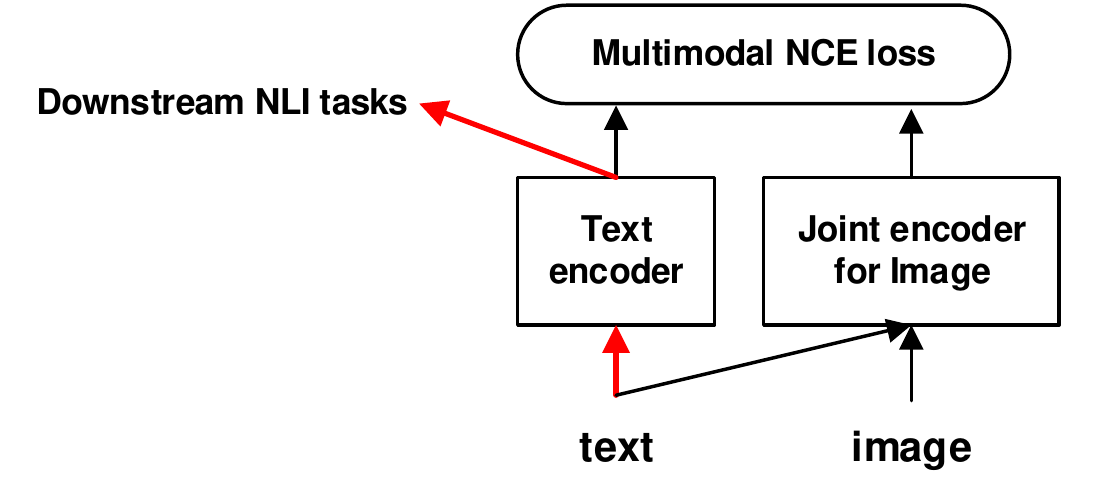}
            \caption{Our proposed multimodal aligned contrastive decoupled network. When adapting to downstream NLI tasks, we directly leverage the representation by the text encoder through the red lines, which only requires text as input.}
        \label{fig:framework:c}
    \end{subfigure}
\caption{Comparison of different multimodal SSL approaches. }
\label{fig:framework}
\end{figure}

The idea of adapting multimodal in SSL is not new. 
According to~\cite{Su2020VL-BERT}, we briefly divide previous multimodal SSL approaches into two categories based on their encoder infrastructures. As shown in Fig.~\ref{fig:framework:a}, the first category uses one joint encoder to represent the multimodal inputs~\cite{sun2019videobert,alberti2019fusion,li2019visualbert,li2019unicoder,Su2020VL-BERT}. Obviously, if the downstream task is only for plain text, we cannot extract the representation of text separately from the joint encoder. So the first category is infeasible for the natural language inference. The second category~\cite{lu2019vilbert,tan2019lxmert,sun2019videobert} first encodes the text and the image separately by two encoders. Then it represents the multimodal information via a joint encoder over the lower layer encoders. This is shown in Fig.~\ref{fig:framework:b}. Although the textual representation can be extracted from the text encoder in the lower layer, such representation does not go through the joint learning module and contains little visual knowledge. In summary, the encoders in previous multimodal SSL approaches are coupled. If only textual inputs are given, they cannot effectively incorporate visual knowledge in their representations. Thus their help for entailing the contradiction between \textcircled{a} and \textcircled{b} is limited.

In order to benefit from multimodal data in plain text inference, we
propose the \underline{M}ultimodal \underline{A}ligned \underline{C}ontrastive \underline{D}ecoupled learning (MACD) network. This is shown in Fig.~\ref{fig:framework:c}. Its text encoder is decoupled, which only takes the plain text as inputs. Thus it can be directly adapted to downstream NLI tasks. Besides, we use multimodal contrastive loss between the text encoder and the image encoder, thereby forcing the text representation to align with the corresponding image. Therefore even if the text encoder in MACD only takes the plain text as input, it still represents visual knowledge. In the downstream plain text inference tasks, without taking images as input, the text encoder of MACD still implicitly incorporating the visual knowledge learned by the multimodal contrastive loss. Note that we do not need a decoupled image encoder in the SSL. So the image encoder in Fig.~\ref{fig:framework:c} in MACD takes texts as inputs to provides a more precise image encoder. We will elaborate this in section~\ref{sec:decoupled}.

\section{Problem Formulation}

We outline the general decoupled SSL process of MACD in section~\ref{sec:decoupled}, and the downstream unsupervised NLI task in section~\ref{sec:unsup}.

\subsection{Decoupled Multimodal SSL}
\label{sec:decoupled}
For pretraining MACD, we use the multimodal training data $\mathcal{D}_{t2i}=\{x_i,y_i\}_{i=1}^N$ with $N$ samples. Each sample $\{x_i,y_i\}$ consists of a pair of text $x_i$ and image $y_i$, which describe the same context. It is straightforward to extend our method to modalities other than texts and images.

MACD learns from $\mathcal{D}_{t2i}$. Since text2image is many-to-many, we use energy-based models to represent their correlations. We first encode $x_i$ and $y_j$ into one pretext-invariant representation space~\cite{misra2019self}. The encoders are denoted by $f(x_i;\theta_f)$  and $g(x_i,y_i;\theta_g)$, respectively. We define the energy function $\sigma(x_i,y_i) : X \times Y \rightarrow \mathbb{R}$ as
\begin{equation}
\label{eqn:decoupled}
\sigma(x_i,y_i)=d(f(x_i;\theta_f),g(x_i,y_i;\theta_g))
\end{equation}
where $f(x_i;\theta_f)$ denotes the text encoder and $g(x_i,y_i;\theta_g)$ denotes the image encoder. $d$ is a non-parametric distance metric (e.g. cosine). In the rest of this paper, we will use $f(x)$ and $g(x,y)$ instead of $f(x;\theta_f)$ and $g(x,y;\theta_g)$ for convenience.

Note that the text encoder $f(x)$ only takes the text as input, while the image encoder $g(x,y)$ takes both the image and the text as input. The higher the value of the energy function $\sigma()$, the higher the probability that $x$ and $y$ are in the same context, and vice versa.
The forms of the encoders have the following advantages:
\begin{itemize}
\item The text encoder $f(x)$ and the image input $y$ are {\it decoupled}. Therefore we represent $x$ separately without knowing $y$. This allows us to use $f(x)$ in the downstream plain text inference.
\item $g(x,y)$ represents the {\it one-to-many} relationship via implicitly introducing the ``predictive sparse coding''~\cite{gregor2010learning}. One image has multiple corresponding texts. To use energy-based models to represent the one-to-many relationship, one common approach is to introduce a noise vector $z$ to allow multiple predictions through one image~\cite{bojanowski2018optimizing}. Note that such $z$ can be quickly estimated by the given text $x$ and image $y$~\cite{gregor2010learning}. In our proposed image encoder $g(x,y)$, although $z$ is not explicitly introduced, the encoder allows multiple predictions for one image via taking different images as input. Besides, it allows the image to interact with the text in the inner computation, which is an implicit alternative for the predictive $z$.
\end{itemize}

\subsection{Downstream Unsupervised NLI}
\label{sec:unsup}

We use the representation from the pre-trained multimodal SSL to predict the relations of natural language sentence pairs under the unsupervised learning scenario. The testing data can be formulated as $\mathcal{D}_{test}=\{x^T_i,z_i\}_{i=1}^M$, each $x^T_i=(x^{1}_i, x^{2}_i)$ is composed of a sentence pair $x^{1}_i$ and $x^{2}_i$. $z_i$ indicates the relation between $x_i^1$ and $x_i^2$. Under the unsupervised setting, we predict $z_i$ for given $x_i^T$ by the similarity of $f(x^1_i)$ and $f(x^2_i)$ (e.g. cosine similarity).

\section{Methods}
This section elaborates our major methodology. In section~\ref{sec:method:global}, we show how we maximize the cross-modal mutual information (MI) for the decoupled representation learning. In section~\ref{sec:structure}, we show how we incorporate the mutual information~(MI) of local structures.
We elaborate the encoders in section~\ref{sec:method:local_encoder}. In Section~\ref{sec:lifelong}, in order to solve the catastrophic forgetting problem, we use lifelong learning regularization to anchor the text.

\subsection{Decoupled Representation Learning by Cross-Modal Mutual Information Maximization}
\label{sec:method:global}

As discussed in section~\ref{sec:intro}, the query object and its context determine the inference. NLI depends on whether the two sentences are in the same context. In this paper, we consider context from different modalities (e.g. text or images). 

Mutual information maximization has become a trend for SSL~\cite{tian2019contrastive,hjelm2018learning}. For cross-modal SSL, we also leverage mutual information $\mathcal{I}(X,Y)$ to represent the correspondence between the text and the image. Intuitively, high mutual information means that the text and the image are well-matched. More formally, the goal of multimodal representation learning is to maximize their mutual information:
\begin{equation}
\label{eqn:e}
\small
\mathcal{I}(X,Y) = \sum_{x,y} P(x,y) \log \frac{P(x|y)}{P(x)}
\end{equation}

Eqn.~\eqref{eqn:e} is intractable and thereby hard to compute. To approximate and maximize $\mathcal{I}(X,Y)$, we use Noise-Contrastive Estimation (NCE)~\cite{gutmann2010noise,oord2018representation}. First, we use the function $\sigma(x,y)$ to represent the term $\frac{P(x|y)}{P(x)}$ in Eqn.~\eqref{eqn:e}:
\begin{equation}
\label{eqn:ptipt}
\small
\sigma_\text{global}(x,y) \propto \frac{P(x|y)}{P(x)}
\end{equation}
where $\sigma_\text{global}(x,y): X \times Y \rightarrow \mathbb{R}$ is not a real probability and can be unnormalized. Here we use the notation ``global'' for the representation learning of a complete text or a complete image to distinguish from the local structures in section~\ref{sec:structure}.

To compute the cross-modal mutual information, we first encode $x$ and $y$ to $f_\text{global}(x)$ and $g_\text{global}(y)$, respectively.
Then we use the similarities of their encodings to model $\frac{P(x|y)}{P(x)}$. Note that $g_\text{global}(y)$ is a specific form of $g(x,y)$ in Eqn.~\eqref{eqn:decoupled}. So $f_\text{global}(x)$ and $g_\text{global}(y)$ satisfy the form of $f$ and $g$ in Eqn.~\eqref{eqn:decoupled}. We will show how to incorporate the linguistic input when designing the encoder of local visual structures in section~\ref{sec:method:local_encoder}. We follow~\cite{misra2019self} to compute the pretext-invariant energy function by the exponential function of their cosine similarity:
\begin{equation}
\label{eqn:energy}
\small
\begin{aligned}
\sigma_\text{global}(x,y) & = d(f_\text{global}(x),g_\text{global}(y)) \\
& = \exp(\frac{cosine(f_\text{global}(x),g_\text{global}(y))}{\tau_\sigma})
\end{aligned}
\end{equation}
where $\tau_\sigma$ is a hyper-parameter of temperature.

To estimate $\sigma_\text{global}(x,y)$ and maximize the mutual information in Eqn.~\eqref{eqn:e}, the NCE loss~\cite{oord2018representation} provides a valid toolkit.
By taking the posterior probability $P(y|x)$, the NCE loss is defined as:
\begin{equation}
\small
\label{eqn:nce:yx}
\begin{aligned}
&\loss^{\text{NCE}:P(y|x)}(X,Y) = -\mathbb{E}_{x,y \sim P(y|x)\tilde{P}(x) }\{ \log \sigma_\text{global}(x,y) \\
&  - \log \sum_{y' \sim P(y)} \sigma_\text{global}(x,y') \}
\end{aligned}
\end{equation}
where $\tilde{P}(x)$ denotes the real distribution of $x$, $P(y|x)\tilde{P}(x)$ denotes the distribution of $y$ for given $x$, and $P(y)$ denotes the noise distribution of $y$. Thus minimizing Eqn.~\eqref{eqn:nce:yx} can be seen as identifying the positive image $y \sim P(y|x)$ for given $x$ from the noise image distribution $y \sim P(y)$.

It has been proved~\cite{oord2018representation} that $\loss^{\text{NCE}:P(y|x)}(X,Y)$ provides the lower bound of $\mathcal{I}(X,Y)$:
\begin{equation}
\small
\mathcal{I}(X,Y)\ge \log N' -\loss^{\text{NCE}:P(y|x)}
\end{equation}
where $N'$ denotes the number of noise samples and can be seen as a constant. So instead of maximizing $\mathcal{I}(X,Y)$ directly, we minimize $L^{\text{NCE}:P(y|x)}(X,Y)$ instead to maximize its lower bound.

Symmetrically, we also compute the NCE loss by taking the posterior probability $P(x|y)$. 
We define $\loss^{\text{NCE}:P(x|y)}$ as:
\begin{equation}
\label{eqn:nce:xy}
\small
\begin{aligned}
\loss^{\text{NCE}:P(x|y)}(X,Y)= & -\mathbb{E}_{x,y \sim P(x|y)\tilde{P}(y) }\{\log \sigma_\text{global}(x,y) \\
& - \log \sum_{x' \in P(x)} \sigma_\text{global}(x',y) \}
\end{aligned}
\end{equation}
Eqn.~\eqref{eqn:nce:xy} can be seen as identifying the positive text $x \sim P(x|y)$ for given $y$ from the noise text distribution $x \sim P(x)$.

By combining Eqn.~\eqref{eqn:nce:yx} and Eqn.~\eqref{eqn:nce:xy}, we derive the loss for global MI maximization
\begin{equation}
\small
\label{eqn:global}
\begin{aligned}
\loss^{\text{NCE}}_\text{global}(X,Y)= &\loss^{\text{NCE}:P(x|y)}(X,Y)  + \loss^{\text{NCE}:P(y|x)}(X,Y)
\end{aligned}
\end{equation}
Here we say the MI is global, because it is over the complete text and the complete images, which are contrary to the local structures in section~\ref{sec:structure}. 

{\bf Negative sampling}
In practice, to compute $L^{\text{NCE}:P(y|x)}(X,Y)$, we need to construct noise samples for positive samples. We use all the $\{x_i,y_i\}$ pairs in the same minibatch from $\mathcal{D}_{t2i}$ as $X,Y$. Each $y_i$ is the positive samples of $x_i$ (i.e. $P(y_i|x_i)=1$). For each $x_i \in X$, the noise $y'$ in Eqn.~\eqref{eqn:nce:yx} are sampled from $Y$.
Likewise, to compute $\loss^{\text{NCE}:P(x|y)}(X,Y)$ in Eqn.~\eqref{eqn:nce:xy}, we treat $x_i$ as the positive sample for $y_i$, and other texts from the same minibatch as the noise samples.

\subsection{MI Optimization for Local Structures}
\label{sec:structure}
In this subsection, we incorporate the local information in multimodal contrastive learning. As demonstrated in DIM~\cite{hjelm2018learning}, local information plays a greater role in self-supervised learning than the global information.

We follow BERT~\cite{devlin2018bert} and DIM to use the words and patches as the local structures for the text and the image, respectively. We maximize the MI between the cross-modal local/global structures. We denote a sentence $x$ with $L$ words as $x^{(1)}\cdots x^{(L)}$, and an image $y$ with $M\times M$ patches as $y^{(1)} \cdots y^{(M^2)}$.

\begin{figure*}[!htb]
        \centering
            \includegraphics[scale=0.4]{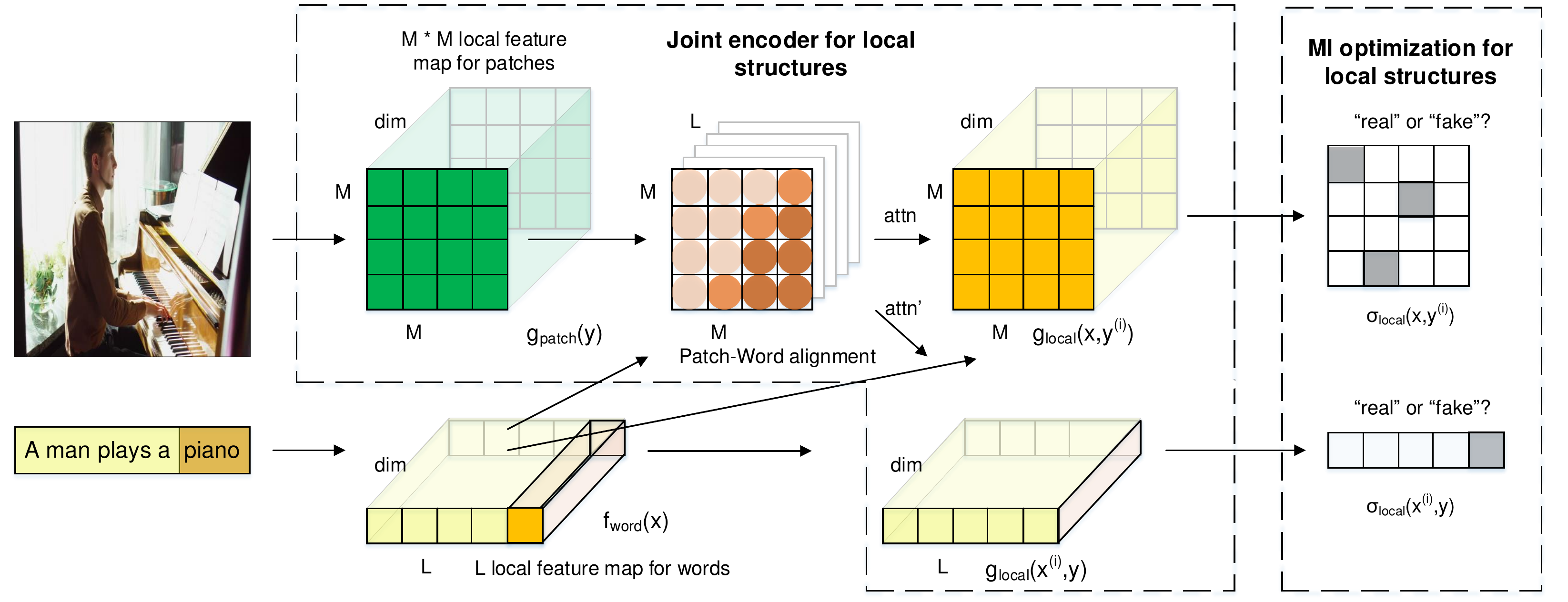}
            \caption{MI maximization for local structures. The local structures for images are joint encoded with text.}
        \vspace{-0.3cm}
        \label{fig:local}
\end{figure*}

Similar to the objective of representation learning of global information, we use NCE as the objective of local information representation learning. The difference is that we use the local structure-based alignment to calculate the energy function, while there is no such objective in the representation learning of global information. This objective allows representation learning to emphasize the alignments of local structures between different modalities, such as the alignment between the word ``piano'' and the corresponding image patches.

Specifically, we use $\loss^{\text{NCE}}_\text{local}(X,Y)$ to represent the loss of local information representation learning. The computation of $\loss^{\text{NCE}}_\text{local}(X,Y)$ follows Eqn.~\eqref{eqn:nce:yx}\eqref{eqn:nce:xy}\eqref{eqn:global}, except that we replace $\sigma_{global}$ with $\sigma_{local}$ based on the local information alignment. We will elaborate on $\sigma_{local}$ in section~\ref{sec:method:local_encoder}.

\subsection{Alignment-based Local Energy Function and Representation Learning}
\label{sec:method:local_encoder}

In this subsection, we show the details of the local energy function $\sigma_{local}$ and the encoders for local structures.

Following the form of Eqn.~\eqref{eqn:decoupled}, we denote the encoders for the local structures of text as $f_\text{word}(x^{(i)})$. We denote the joint encoder for patches as $g_\text{local}(x,y^{(i)})$, which represents the linguistic information of patch $y^{(i)}$. Note that the encoder $f_\text{word}(x^{(i)})$ is still decoupled and represents the local linguistic structures without taking image as input. On the other hand, the encoder $g_\text{local}(x,y^{(i)})$ for the local visual structure explicitly incorporate the linguistic information, which is more precise due to the discussion in section~\ref{sec:decoupled}.

For a sentence $x$ with $L$ words $x^{(1)}\cdots x^{(L)}$, we represent its local information by encoding it into a local feature map $\small{\bm{f}_\text{word}(x)=\begin{pmatrix}f_\text{word}(x^{(1)}) \cdots f_\text{word}(x^{(L)})\end{pmatrix} \in \mathbb{R}^{dim \times L}}$. For an image $y$ with $M\times M$ patches $y^{(1)} \dots y^{(M^2)}$, we represent its spatial locality by encoding it into a feature map $\bm{g}_\text{patch}(y)=\begin{pmatrix}g_\text{patch}(y^{(1)}) \cdots g_\text{patch}(y^{(M^2)})\end{pmatrix}$. 

The local information across modalities has obvious correlation characteristics~\cite{xu2018attngan}. For example, a word is only related to some patches of the image, but not to other patches. As shown in Fig.~\ref{fig:framework:c}, our proposed image encoder is coupled with the text representation. Therefore we assign the local structures with different weights to achieve a more precise image encoder. This is achieved by the attention mechanism in the joint encoder:
\begin{equation}
\small
\begin{aligned}
&g_\text{local}(x,y^{(i)}) =\frac{exp(attn_{i,j}/\tau_c)}{\sum_k exp(attn_{k,i}/\tau_c)} \bm{f}_\text{word}(x)
\end{aligned}
\end{equation}
where $\tau_c$ denotes the temperature, $attn_{i,j}$ denotes the attention of the $i$-th word to the $j$-th patch:
\begin{equation}
\small
attn_{i,j}=\frac{exp(f_\text{word}(x^{(i)})^{\top} g_\text{patch}(y^{(j)}))}{\sum_k exp(f_\text{word}(x^{(i)})^{\top} g_\text{patch}(y^{(k)}))}
\end{equation}

We compute the alignment score for the local textual structures by:
\begin{equation}
\begin{aligned}
&\sigma_\text{local}(x,y^{(i)}) = d(g_\text{patch}(y^{(i)}),g_\text{local}(x,y^{(i)}))
\end{aligned}
\end{equation}
Here we abuse the notation of $\sigma_\text{local}$ since we will use $\sigma_\text{local}(x,y^{(i)})$ to compute $\sigma_\text{local}(x,y)$. 

Symmetrically, we also compute the alignment score for the local visual structures by
\begin{equation}
\small
\begin{aligned}
& attn'_{i,j}=\frac{exp(f_\text{word}(x^{(i)})^{\top} g_\text{patch}(y^{(j)}))}{\sum_k exp(f_\text{word}(x^{(k)})^{\top} g_\text{patch}(y^{(j)}))}\\
& g_\text{local}(x^{(i)},y) =\frac{exp(attn'_{i,j}/\tau_c)}{\sum_k exp(attn'_{i,k}/\tau_c)} \bm{g}_\text{patch}(y)\\
&\sigma_\text{local}(x^{(i)},y) = d(f_\text{word}(x^{(i)}), g_\text{local}(x^{(i)},y))\\
\end{aligned}
\end{equation}


We compute the energy function of $x$ and $y$ based on local structure alignments by:
\begin{equation}
\begin{aligned}
\sigma_\text{local}(x,y) = & \log \sum_{i=1}^L exp(\sigma_\text{local}(x^{(i)},y)) \\
& + \log \sum_{i=1}^{M^2} exp(\sigma_\text{local}(x,y^{(i)}))
\end{aligned}
\end{equation}
How the model uses the attention mechanism to represent the interactions among local structures and how the energy function is computed is shown in Fig.~\ref{fig:local}.

\subsection{Anchor Text via Lifelong Learning}
\label{sec:lifelong}
In this subsection, we illustrate how to solve the catastrophic forgetting problem by the lifelong learning regularization.

If we only use the loss in Eqn.~\eqref{eqn:global}, the text encoder $f(x;\theta_f)$ will tend to only learn vision-related features for text. Since our downstream problem is over the plain text, NLI still relies more on textual features instead of visual features. Compared with the single modality unsupervised natural language representation learning~\cite{devlin2018bert}, the multimodal model will even perform worse. Similar phenomena called catastrophic forgetting or negative transfer~\cite{sun2020sparsing} often occurs in multi-task learning.

To avoid the catastrophic forgetting, we keep the model's representation for general text while ensuring that it learns visual features.
More generally, since there are only data of a certain modality (i.e. plain text) in the downstream task,
we anchor this modality in the multimodal SSL phase.
We add lifelong learning regularization~\cite{li2017learning} to achieve modality anchoring. For the text encoder, we keep its original textual representation (e.g. by masked language model (MLM) and next sentence prediction in BERT) while learning new visual knowledge. To do this, we follow~\cite{li2017learning} and introduce the distance from the existing text encoder to the original text encoder as the training loss. 

Specifically, we use BERT~\cite{devlin2018bert} to initialize our text encoder $f(x)$. During multimodal SSL, we keep the textual representation consistent with the original BERT. According to the ablation study in DistilBERT~\cite{sanh2019distilbert}, we use the knowledge distillation loss~\cite{44873} and cosine loss as regularization:
\begin{equation}
\begin{aligned}
&\loss_{anchor}(X)=\mathbb{E}_{x\sim \tilde{P}(x)} [ \\
&-\epsilon\sum_{i=1}^{dim} \frac{f_i(x)^{1/\tau'}}{\sum_j f_j(x)^{1/\tau'}} \log \frac{f'_i(x)^{1/\tau'}}{\sum_j f'_j(x)^{1/\tau'}} \\
&-(1-\epsilon)cosine(f(x),f'(x))]
\end{aligned}
\end{equation}
where $f'(x)$ denotes the textual representation by the original BERT encoder, $f_i(x)$ denotes the $i$-th dimension of $f(x)$, and $\tau'$ is the temperature.

By combing the lifelong learning regularization, we obtain the final loss for SSL:
\begin{equation}
\small
\begin{aligned}
\hat{\theta}_x, \hat{\theta}_y, \hat{\theta}_\alpha &= \argmax_{\theta_f,\theta_g} \gamma \loss^\text{NCE}_\text{global}(X,Y) \\
 & + \beta \loss^\text{NCE}_\text{local}(X,Y) +
 (1-\gamma-\beta) \loss_{anchor}(X)
\end{aligned}
\end{equation}


\section{Experiments}
\subsection{Setup}
\label{sec:expsetup}
All the experiments run over a computer with 4 Nvdia Tesla V100 GPUs.

{\bf Datasets} We use Flickr30k~\cite{young2014image} and COCO~\cite{lin2014microsoft} as the text2image dataset $\mathcal{D}_{t2i}$ for self-supervised learning. We use STS-B~\cite{cer2017semeval} and SNLI~\cite{snli:emnlp2015} as the downstream NLI tasks for evaluation. STS-B is a collection of sentence pairs, each of which has a human-annotated similarity score from 1 to 5. The task is to predict these scores. We follow GLUE~\cite{wang2018glue} and use Pearson and Spearman correlation coefficients as metrics. SNLI is a collection of human-written English sentence pairs, with manually labeled categories {\it entailment}, {\it contradiction}, and {\it neutral}.
Note that for STS-B, some sentence pairs drawn from image captions overlap with Flickr30k. So in order to avoid the potential information leak,
we remove all sentence pairs drawn from image captions in STS-B to construct a new dataset {\it STS-B-filter}. Similarly, we remove all sentence pairs in SNLI whose corresponding images occur in the training split of $\mathcal{D}_{t2i}$ to construct {\it SNLI-filter}.

The statistics of these datasets are shown in Table~\ref{tab:datasets}. In addition, Flickr30k has 22248 images for training, 9535 images for development. COCO has 82783 images for training, 40504 images for development.

\begin{table}[!htb]
\small
\centering
\setlength{\tabcolsep}{2.2pt}
\vspace{-0.4cm}
\begin{tabular}{l|l|lll}
 \hline
          & Type                & \multicolumn{3}{c}{\#Text}  \\
          &                     & Train   & Dev     & Test     \\  \hline \hline
Flickr30k & Text2Image          & 111240  & 47675   & -           \\
COCO      & Text2Image          & 414113  & 202654  & -       \\ \hline
STS-B     & Text Similarity & 5749    & 1500    & 1379        \\
STS-B-filter     & Text Similarity & 3749    & 875    & 754    \\
SNLI      & NLI                 & 549367  & 9842    & 9824  \\
SNLI-filter      & NLI                 & 157284  & 3321    & 3207  \\  \hline
\end{tabular}
\caption{Statistics of datasets.}
\label{tab:datasets}
\vspace{-0.6cm}
\end{table}

\subsection{Model Details}
\label{sec:modeldetails}
\nop{
{\bf Encoder details} We use BERT as the text encoders $f(x;\theta_x)$ and $f_\text{local}(x;\theta_x)$. The local information $f_\text{local}(x^{(i)};\theta_x)$ is the feature vector of the $i$-th word through BERT. The global information $f(x;\theta_x)$ is the embedding of the $[CLS]$ token of the sentence.
We use Inception-v3~\cite{szegedy2016rethinking} pre-trained on ImageNet~\cite{russakovsky2015imagenet} as the image encoder $g(y;\theta_y)$ and $g_\text{local}(y;\theta_y)$. We use Inception-v3 because its inner ``mixed 6e'' layer is also used for the auxiliary image classification task while training. So the vectors of that a layer is more applicable to represent the patches. We extract the local features from the ``mixed 6e'' layer of Inception-v3 as $g_\text{local}(y;\theta_y)$ to represent the patches, which will yield $17 \times 17$ patches. The global feature $g(y;\theta_y)$ is extracted from the last average pooling layer of Inception-v3. To guarantee that the image encoder and the text encoder are in the same space, we project the feature vectors of the image encoder to the dimension of 768, which is the dimension of the text encoder.
}

{\bf Encoder details} We use BERT-base as the text encoder $f_\text{global}$. The local information $f_\text{word}(x^{(i)})$ is the feature vector of the $i$-th word through BERT. We use {\it Resnet-50} as the image encoder $g_\text{global}$. We use the encoding before the final pooling layer as the representations of $M^2$ patches $g_\text{patch}(y^{(i)})$. To guarantee that the image encoder and the text encoder are in the same space, we project the feature vectors of the image encoder to the dimension of 768, which is the dimension of BERT.

{\bf Unsupervised NLI} We compute the similarity of two sentences via the cosine of their representations learned by MACD. For STS-B, such similarities are directly used to compute the Pearson and Spearman correlation coefficients. For SNLI, we make inferences based on whether the similarity reaches a certain threshold. More specifically, if the similarity $>=\psi_1$, we predict ``entailment''. If the similarity $<\psi_2$, we predict ``contradiction''. Otherwise we predict ``neutral''.


{\bf Competitors} We compare MACD with the single-modal pre-training model BERT, and multimodal pre-training model LXMERT~\cite{tan2019lxmert} and VilBert~\cite{lu2019vilbert}. Both LXMERT and VilBert use the network architecture as in Fig.~\ref{fig:framework:b}. We extract the lower layer text encoder for unsupervised representation and fine-tuning. We also compare MACD with classical NLP models, including BiLSTM and BiLSTM+ELMO~\cite{Peters:2018}.


{\bf Hyper-parameters} We list the hyper-parameters below. For $\psi_1$ and $\psi_2$, we use the best set of values chosen in the grid search from range $\{-1,-0.95,-0.9,\cdots,1\}$. For $\tau_\sigma$ and $\tau_c$, we use the best set of values chosen in the grid search from range $\{0.01, 0.1, 1\}$. For $\tau'$, $\epsilon$, $\gamma$ and $\beta$, we follow their settings in DistilBert~\cite{sanh2019distilbert}.

\begin{table}[!htb]
\small
\centering
\setlength{\tabcolsep}{2.2pt}
\begin{tabular}{l|l|l|l|l|l}
\hline
$\tau_\sigma$ &  $\tau_c$ & $\tau'$   & $\epsilon$ & $\gamma$ & $\beta$ \\ \hline
0.1                                          & 1                    & 2                     & 5/6                     & 1/3                   & 1/3                  \\ \hline \hline
Batch Size                 & lr                         & Epochs                 & Grad Acc &     $\psi_1$                    &     $\psi_2$                  \\ \hline
64                         & 1e-4                   & 10                     & 8                     &    0.80                     &     0.55\\ \hline
\end{tabular}
\caption{Hyper-parameters for self-supervised learning. ``lr'' means learning rate.}
\vspace{-0.6cm}
\end{table}

\subsection{Main Results}
\label{sec:exp:main}
We evaluate MACD by unsupervised NLI. Table~\ref{tab:uSTS} shows the results on STS-B. MACD achieves significantly higher effectiveness than single-modal pre-trained model BERT and multimodal pre-trained model LXMERT and VilBert. Note that LXMERT and VilBert use more text2image corpora for pre-training than MACD. This verifies that the joint encoder in previous multimodal SSL cannot represent visual knowledge well in their text encoder. So their adaptations to the single-modal problem are limited.


To our surprise, the unsupervised MACD even outperforms fully-supervised models such as BiLSTM and BiLSTM+ELMO. Here the results of BiLSTM and BiLSTM+ELMO for STS-B are directly derived from GLUE~\cite{wang2018glue}. This verifies the effectiveness of MACD.

\begin{table}[!htb]
\small
\centering
\begin{tabular}{l|ll|ll}
 \hline
                               & \multicolumn{2}{c|}{STS-B}  & \multicolumn{2}{c}{STS-B-filter}   \\
                               & P.       & S.  & P.  & S.     \\ \hline \hline
BiLSTM (sup.)            & 66.0          & 62.8    &  47.0     &   43.2   \\
BiLSTM+ELMO (sup.)       & 64.0          & 60.2    &  33.3     &   30.7   \\ \hline
BERT        & 1.7           & 6.4            &  5.5   & 12.5     \\
LXMERT        & 42.7           &   47.2      &  35.9  & 40.0    \\
VilBert       &  55.8          &    57.1     &  45.9  & 46.3    \\ \hline
MACD + COCO & 70.1  & 70.2 & 55.1 & 52.4 \\
MACD + Flickr30k  & {\bf 71.5}  &  {\bf 72.1}  & {\bf 55.8}  & {\bf 54.8}   \\ \hline
\end{tabular}
\caption{Effectiveness of unsupervised learning on STS. Baselines with ``(sup.)'' mean they are trained by supervised labels. Other methods are unsupervised. ``P.'' and ``S.'' mean Pearson and Spearman correlation coefficients, respectively.}
\label{tab:uSTS}
\vspace{-0.4cm}
\end{table}

\begin{table}[!htb]
\centering
\small
\begin{tabular}{l|c|c}
 \hline
                                        & SNLI  & SNLI-filter \\
                                        & Acc     & Acc         \\ \hline \hline
BERT                & 35.09          &  35.45   \\
LXMERT             & 39.03        &  40.29    \\
VilBert            &  43.13        &  43.83    \\ \hline
MACD + COCO & {\bf 52.63}  & 53.15 \\
MACD + Filckr30k           & 52.27            &  {\bf 53.20} \\ \hline
\end{tabular}
\caption{Effectiveness on SNLI. All approaches are unsupervised.}
\label{tab:uSNLI}
\vspace{-0.4cm}
\end{table}

We also report the results of MACD on SNLI under the unsupervised setting in Table~\ref{tab:uSNLI}. MACD outperforms its competitors by a large margin.
This verifies the effectiveness of our approach for unsupervised NLI. The experimental results suggest that {\it we achieve natural language inference via multimodal self-supervised learning without any supervised inference labels.} Since MACD+Filckr30k performs better than MACD+COCO in most cases, we will only evaluate MACD+Filckr30k in the rest experiments.

\begin{figure}[!htb]
    \begin{subfigure}[b]{0.23\textwidth}
        \centering
            \includegraphics[scale=0.28]{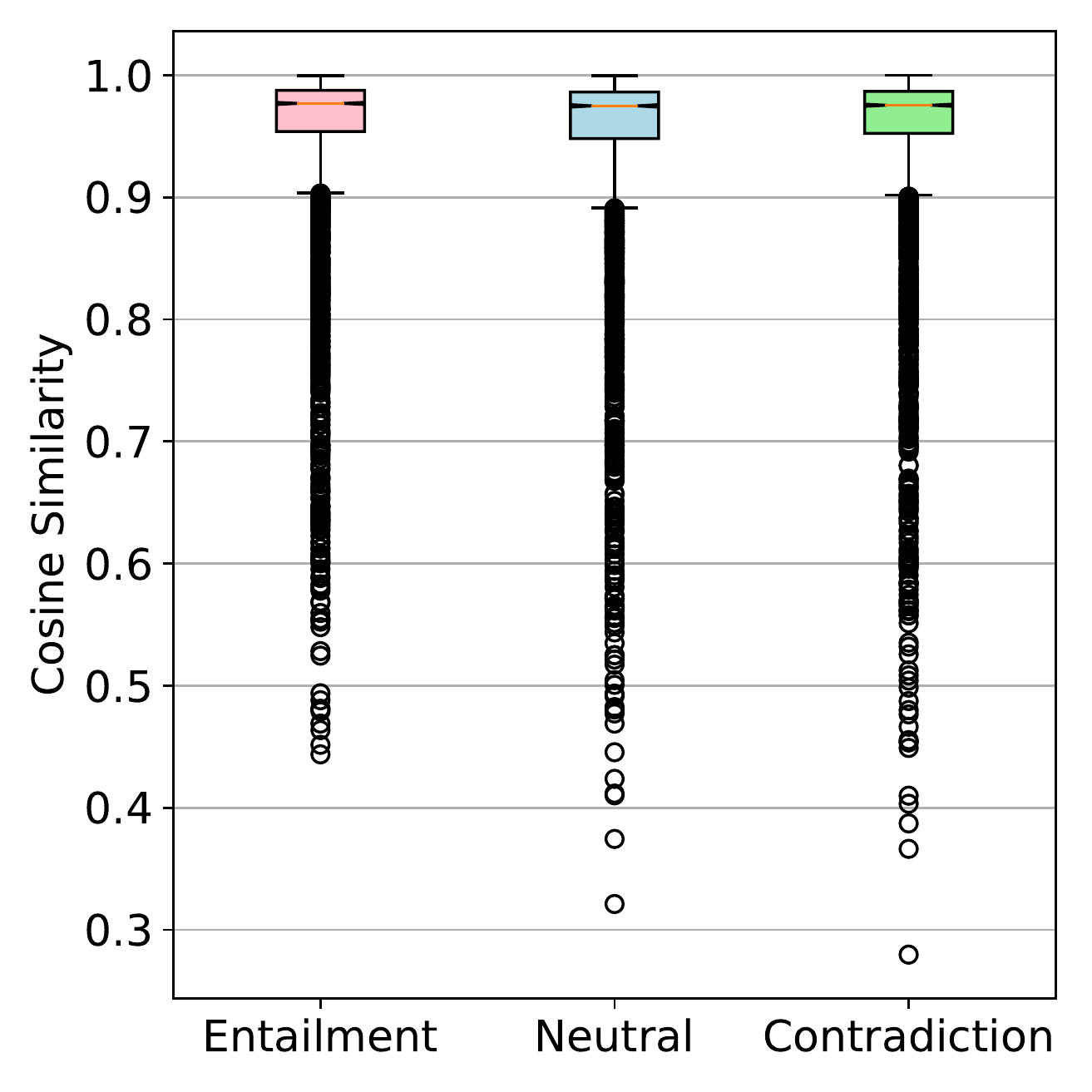}
            \caption{BERT}
            \end{subfigure}
            \hspace{0.1cm}
    \begin{subfigure}[b]{0.23\textwidth}
        \centering
            \includegraphics[scale=0.28]{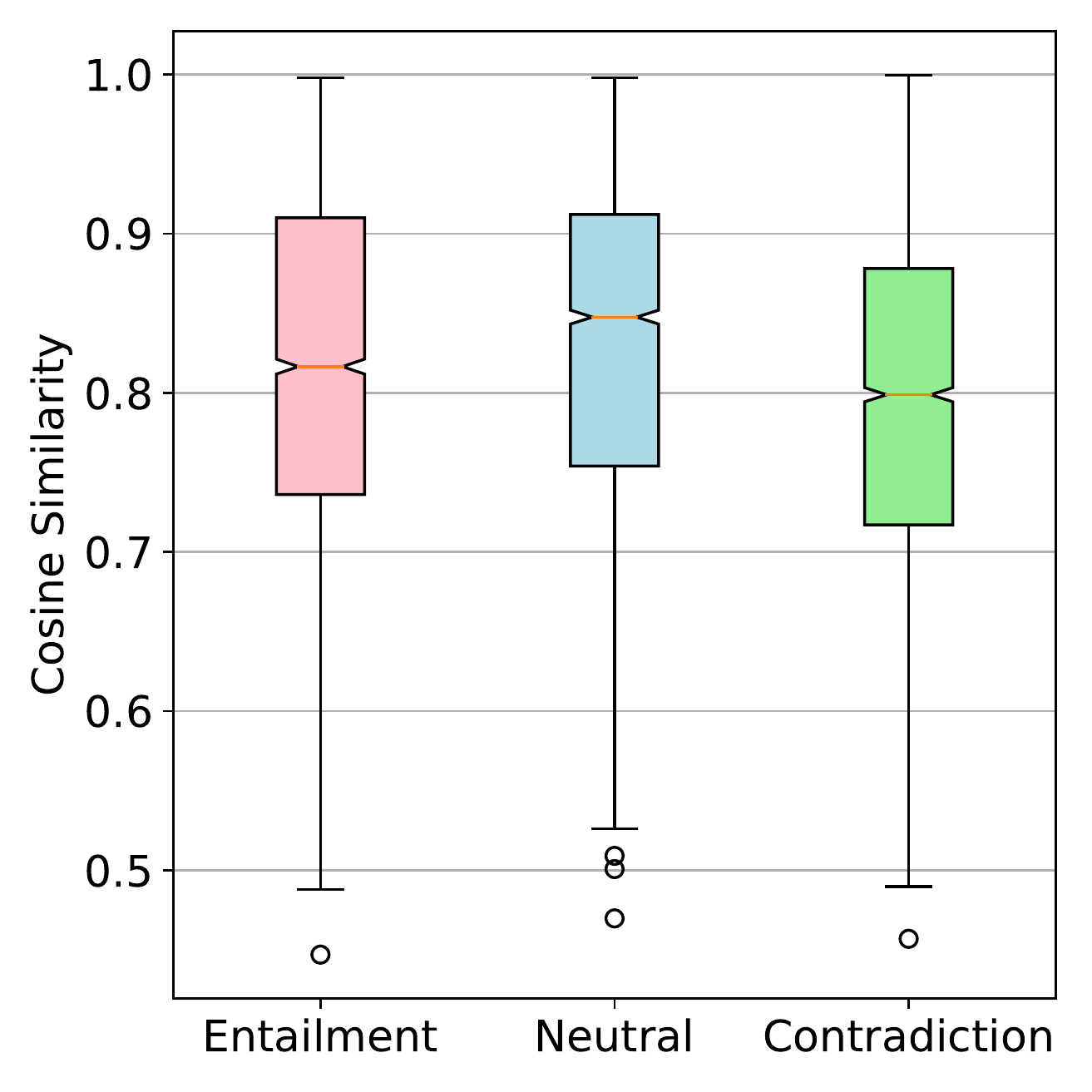}
            \caption{LXMERT}
            \end{subfigure}
    \\
    \begin{subfigure}[b]{0.23\textwidth}
        \centering
            \includegraphics[scale=0.28]{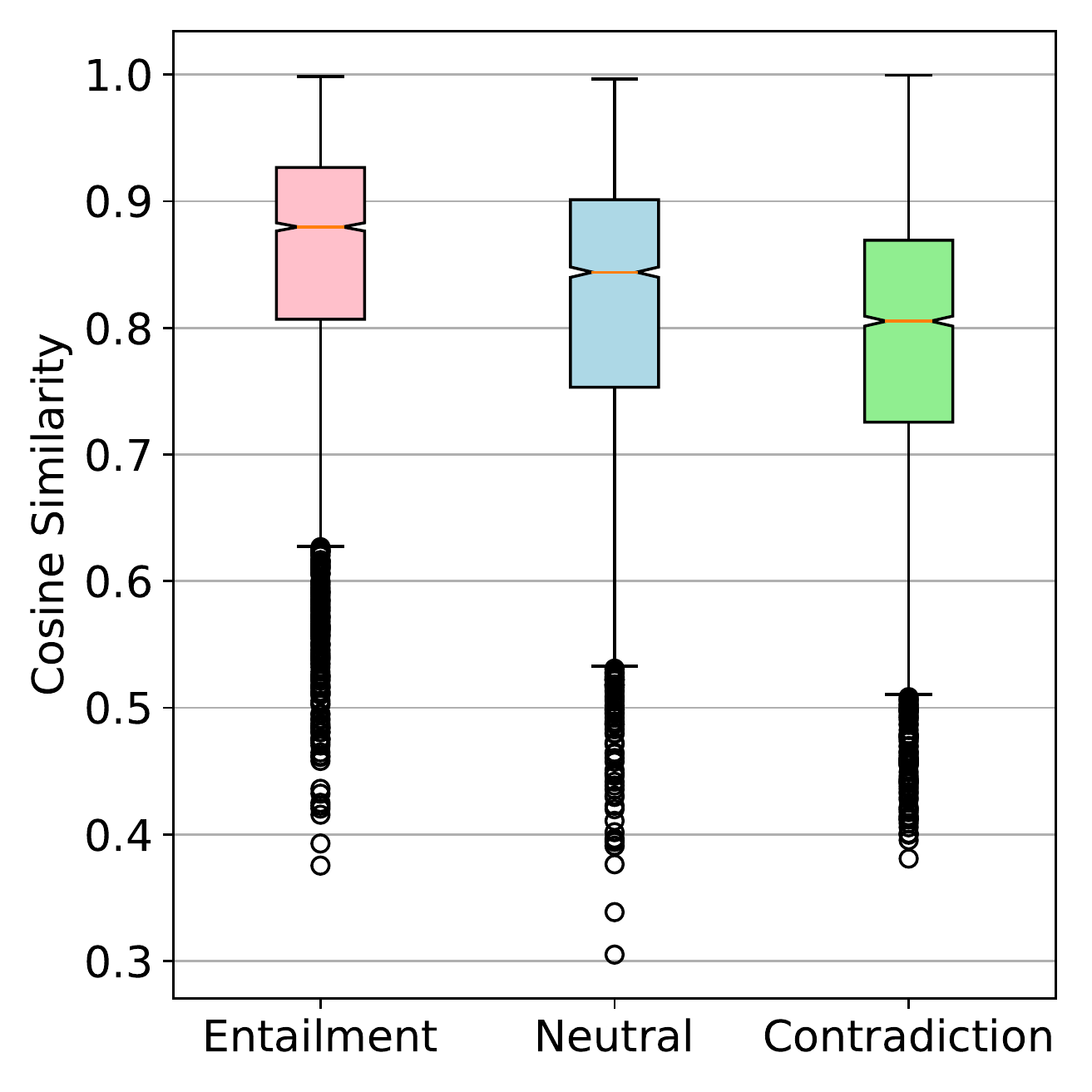}
            \caption{VilBert}
            \end{subfigure}
            \hspace{0.1cm}
    \begin{subfigure}[b]{0.23\textwidth}
        \centering
            \includegraphics[scale=0.28]{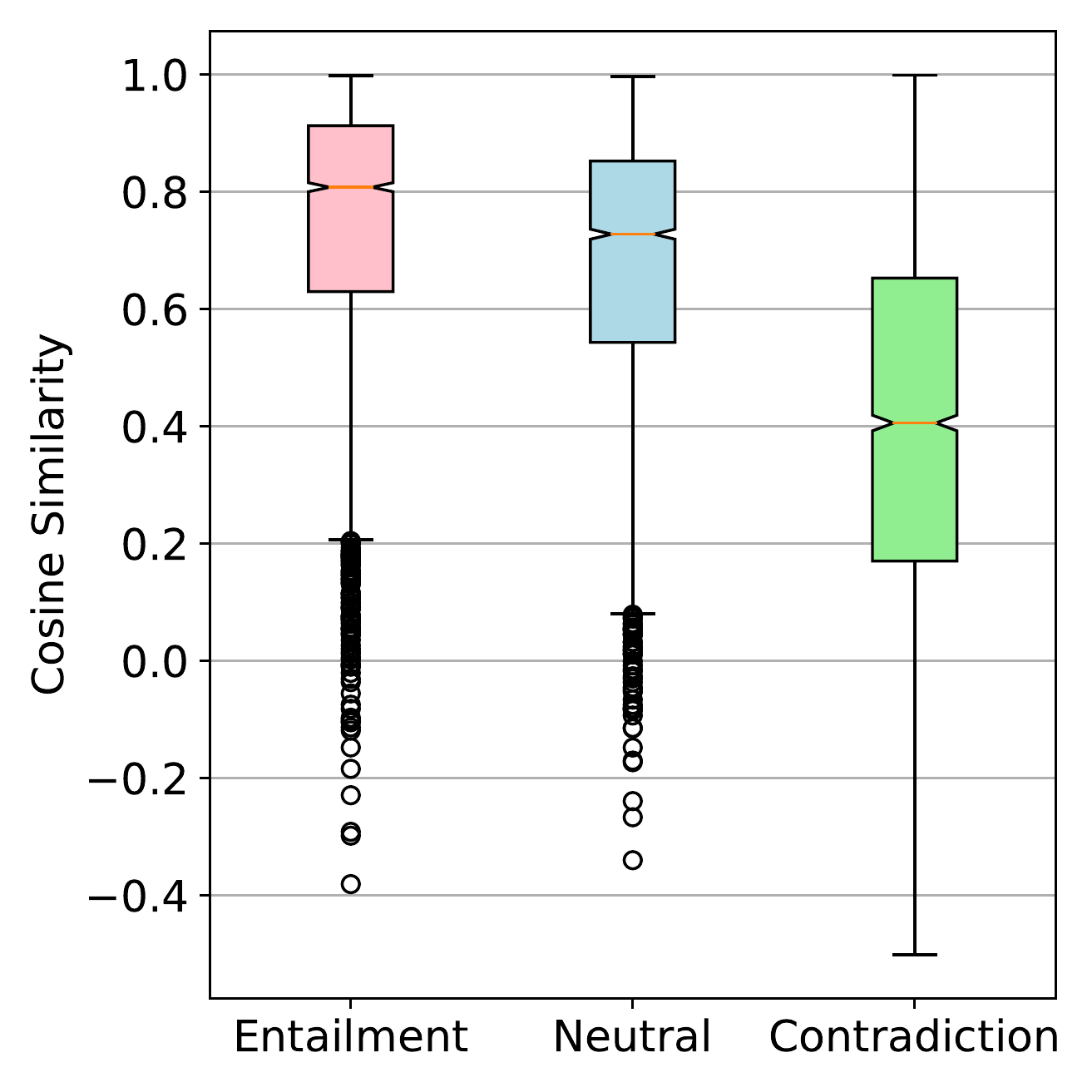}
            \caption{MACD}
            \end{subfigure}
\caption{Categorial distribution visualization.}
\label{fig:boxplot}
\end{figure}

We visualize the distribution of the cosine similarities for samples of different labels in SNLI in Fig.~\ref{fig:boxplot} by boxplot. We found obvious distribution patterns by MACD. In contrast, the distributions of other pre-training models have lower correlations with NLI labels.

\subsection{Fine-tuning}
We also evaluated the effectiveness of MACD when fine-tuned under the semi-supervised learning setting. More specifically, we first initialize the parameters of the text encoder as in MACD, then fine-tune it by the supervised training samples of the downstream tasks. The results are shown in Table~\ref{tab:semisupervised}. MACD also outperforms other approaches. For example, for SNLI-filter, the accuracy of MACD increases by 0.97 compared to the best competitor (i.e. BERT). Note that MACD is the only multimodal method that performs better than BERT. Other multimodal approaches (i.e. LXMERT and VilBert) perform even worse than the original BERT, although they also initialize their text encoders by BERT, and  use more text2image data for SSL than MACD. This verifies the effectiveness of the proposed decoupled contrastive learning model.
\begin{table}[!htb]
\centering
\small
\setlength{\tabcolsep}{1.85pt}
\begin{tabular}{l|c|c|c|c}
\hline
                             & STS-B & STS-B-filter & SNLI & SNLI-filter \\
                             & P. / S.    & P. / S. &  Acc & Acc                   \\ \hline \hline
BERT       & 85.0/83.6    & 75.8/74.6 &  89.37 & 87.15                \\
LXMERT     & 63.3/59.2    & 37.3/28.3 &  87.80 & 83.57 \\
VilBert    & 78.8/77.2    & 63.9/62.2 &  88.49 & 85.69 \\ \hline
MACD       & {\bf 87.1/86.4}   & {\bf 79.5/78.0} & {\bf 90.01} & {\bf 88.12} \\ \hline
\end{tabular}
\caption{Effectiveness of fine-tuning over STS-B and SNLI. ``P.'' and ``S.'' mean Pearson and Spearman correlation coefficients, respectively.}
\label{tab:semisupervised}
\end{table}

To further verify the natural language representation learned by the self-supervised learning and get rid of the influence of its neural network architecture (i.e., BERT), \citet{hjelm2018learning} suggest training models directly over the features learned by SSL. By following its settings~\cite{hjelm2018learning}, we use a linear classifier (SVM) and a non-linear classifier (a single layer perception neural network, marked as SLP) over the features by SSL. The results are shown in Table~\ref{tab:notextencoder}.

\begin{table}[!htb]
\small
\centering
\setlength{\tabcolsep}{1.85pt}
\begin{tabular}{l|c|c|c|c}
\hline
                                                      & STS-B  & STS-B-filter     & SNLI  & SNLI-filter \\
                                                      & P. / S.  & P. / S.    &  Acc     & Acc                 \\ \hline \hline
SVM+BERT                             & 69.8 / 68.3  & 57.1 / 53.3         & 58.77 & 58.87                  \\
SVM+LXMERT                           & 33.0 / 31.3  & 10.2 / 13.2    &  52.28  &  50.98 \\
SVM+VilBert                            & 52.4 / 50.0     & 36.7 / 35.9     & 55.93 & 55.22                  \\
SVM+MACD                               & {\bf 70.0 / 68.4  }   & {\bf 62.2 / 59.3 } & {\bf  61.64 }& {\bf 62.58 }      \\ \hline
SLP+BERT       & 56.2 / 53.5          & 47.3 / 42.0    &  55.07 & 54.19            \\
SLP+LXMERT     & 36.5 / 33.4          & 16.1 / 12.3   &  52.41 & 50.42 \\
SLP+VilBert    & 49.6 / 46.0          & 29.1 / 26.5   &  54.86 & 51.82 \\
SLP+MACD       & {\bf 72.3 / 69.7  }   & {\bf 63.4 / 59.5 }  & {\bf  61.31} & {\bf 60.80}  \\ \hline
\end{tabular}
\caption{Effectiveness of the learned representations.}
\label{tab:notextencoder}
\vspace{-0.4cm}
\end{table}

MACD outperforms the competitors by a large margin. Similar to the results in Table~\ref{tab:semisupervised}, although MACD, LXMERT, and VilBert are all trained by multimodal data, only MACD performs better than the original text encoder (i.e. BERT).

\subsection{Ablations}
In addition to the decoupled contrastive learning model, we propose two optimizations by adding the local structures into account, and by regularizing the model on the text mode via lifelong learning. In order to verify the effectiveness of the two optimizations, we compare MACD with its ablations. The results of unsupervised NLI are shown in Table~\ref{tab:ablation_unsup}. The results show that the effectiveness decreases when the proposed optimizations are removed.

\begin{table}[!htb]
\small
\centering
\begin{tabular}{ l|c|c }
\hline
                               & STS-B  & STS-B-filter                         \\
                               & P. / S.  & P. / S.          \\ \hline \hline
MACD      & \textbf{71.5} / \textbf{72.1} & {\bf 55.8} / {\bf 54.8}          \\
\text{  } -local        & 71.0 / 70.9  & 55.0 / 52.6           \\
\text{  } -lifelong     & 70.7 / 70.8  & 54.9 / 52.3           \\
\text{  } -local -lifelong &  69.6 / 69.7     & 53.0 / 52.0          \\ \hline
\end{tabular}
\caption{Ablations.}
\label{tab:ablation_unsup}
\vspace{-0.6cm}
\end{table}

\subsection{Case studies: Nearest-neighbor analysis}
To give a deeper insight into the learned representation, we analyze the $k$ nearest neighbors over the representations. For the query sentence randomly sampled from Flickr30k, we show the results of the 3 nearest sentences according to their L1 distances in Table~\ref{tab:nearestneighbor}. The results of MACD are more interpretable than BERT.

\begin{table}[!htb]
\small
\setlength\extrarowheight{2pt}
\begin{tabularx}{\columnwidth}{ c|X }
 \hline
Query     & Someone is wearing a large white dress in a crowd.                      \\ \hline
MACD No.1  & Lady dressed in white on blanket in middle of crowd.                                                 \\
MACD No.2  & Women in white robes, dancing with half their face painted.                                      \\
MACD No.3  &  A group of women dressed in white are dancing in the street.                                  \\ \hline
BERT No.1 & A man is standing alone in a boat.       \\
BERT No.2 & A bald man is standing in a crowd.              \\
BERT No.3 & A woman is taking a picture of a man.                               \\ \hline
\end{tabularx}
\caption{Nearest-neighbor on the encoded text.}
\label{tab:nearestneighbor}
\vspace{-0.6cm}
\end{table}

\section{Conclusion}

In this paper, we study the multimodal self-supervised learning for unsupervised NLI. 
The major flaw of previous multimodal SSL methods is that they use a joint encoder for representing the cross-modal correlations. This prevents us from integrating visual knowledge into the text encoder. We propose the multimodal aligned contrastive decoupled learning (MACD), which learns to represent visual knowledge while using only texts as inputs. In the experiments, our proposed approach steadily surpassed other methods by a large margin.

\newpage

{\small
\section*{Acknowledgments}
This paper was supported by National Natural Science Foundation of China (No. 61906116, No. 61732004), by Shanghai Sailing Program (No. 19YF1414700).
}

\bibliographystyle{acl_natbib}
\bibliography{ssl-nli}

\end{document}